\documentclass{article}

\usepackage{mlmath}

    \usepackage[preprint]{neurips_2022}

\usepackage[utf8]{inputenc} 
\usepackage[T1]{fontenc}    
\usepackage{hyperref}       
\usepackage{url}            
\usepackage{booktabs}       
\usepackage{amsfonts}       
\usepackage{nicefrac}       
\usepackage{microtype}      
\usepackage{xcolor}         
\usepackage{multirow}
\usepackage{ulem}
\usepackage{soul}
\usepackage{tabularx}
\newcommand{\tabincell}[2]{\begin{tabular}{@{}#1@{}}#2\end{tabular}}

\usepackage{times}
\usepackage{graphicx}
\usepackage{subfigure}
\usepackage{caption}
\usepackage{amsmath,amsfonts,amssymb,mathtools,amsthm}

\newtheorem{thm}{Theorem}

\newtheorem{prop}{Proposition}

\newtheorem*{theorem*}{Theorem}

\newtheorem*{lemma*}{Lemma}

\newtheorem*{property*}{Property}
\newtheorem*{remark}{Remark}
\newtheorem{definition}{Definition}

\newtheorem{assumption}{Assumption}
\newtheorem*{assumption*}{Assumption}
\newtheorem*{proposition*}{Proposition}
\newtheorem{setting}{Setting}
\newtheorem*{setting*}{Setting}

\title{Dropout in training neural networks: flatness of solution and noise structure}

\author{%
Zhongwang Zhang, Hanxu Zhou \& Zhi-Qin John Xu \thanks{Corresponding author: xuzhiqin@sjtu.edu.cn.} \\
School of Mathematical Sciences, Institute of Natural Sciences, MOE-LSC and \\
Qing Yuan Research Institute, Shanghai Jiao Tong University\\
\texttt{\{0123zzw666, zhouhanxu, xuzhiqin\}@sjtu.edu.cn} \\
}

\begin{document}

\maketitle

\begin{abstract}
It is important to understand how the popular regularization method dropout helps the neural network training find a good generalization solution. In this work, we show that the training with dropout finds the neural network with a flatter minimum compared with standard gradient descent training. We further find that the variance of a noise induced by the dropout is larger at the sharper direction of the loss landscape and the Hessian of the loss landscape at the found minima aligns with the noise covariance matrix by experiments on various datasets, i.e., MNIST, CIFAR-10, CIFAR-100 and Multi30k, and various structures, i.e., fully-connected networks, large residual convolutional networks and transformer. For networks with piece-wise linear activation function and the dropout is only at the last hidden layer, we then theoretically derive the Hessian and the covariance of dropout randomness, where these two quantities are very similar. This similarity may be a key reason accounting for the goodness of dropout.  
\end{abstract}

\section{Introduction}
Dropout is used with gradient-descent-based algorithms for training DNNs \citep{hinton2012improving,srivastava2014dropout}, \textcolor{black}{which drives the state-of-the-art test performance in deep learning \citep{tan2019efficientnet,helmbold2015inductive}.} During training, the output of each neuron is multiplied with a random variable with probability $p$ as $1/p$ and $1-p$ as zero. Note that $p$ is called dropout rate, and every time for computing concerned quantity, the variable is randomly sampled at each feedforward operation. Dropout has been an indispensable trick in the training of deep neural networks (DNNs).

\textcolor{black}{The noise structure in the training dynamics is important. For example, the noise structure of SGD helps find a flat solution  \citep{keskar2016large,feng2021inverse,zhu2018anisotropic}.} Similar to SGD, training with dropout is equivalent to that with some specific noise. \textcolor{black}{The implicit regularization behind this specific noise structure finds solutions with better generalization  \citep{hinton2012improving,srivastava2014dropout,wei2020implicit}. }


To understand what kind of noise benefits the generalization of training, \textcolor{black}{in this work, }we first study the characteristic of the minima found \textcolor{black}{with the dropout regularization}. We show that compared with the standard gradient descent (GD), the GD with dropout selects flatter minima. As suggested by many existing works \citep{keskar2016large,neyshabur2017exploring,zhu2018anisotropic}, flatter minima are more likely to have better generalization and stability. 

To explain why dropout can find flat minima, we then explore the relation between the flatness of the loss landscape and the noise structure induced by dropout at minima through three methods and obtain consistent results as follows: i) Inverse variance-flatness relation: The noise is larger at the sharper direction of the loss landscape; ii) Hessian-variance alignment relation: The Hessian of the loss landscape at the found minima aligns with the noise covariance matrix.

These two relations are intuitively consistent and may help the training select flatter minima.  Our experiments are conducted over several representative datasets, i.e., MNIST \citep{lecun1998gradient}, CIFAR-100 \citep{krizhevsky2009learning} and Multi30k \citep{elliott2016multi30k}, and network structures, i.e., fully-connected neural networks, ResNet-20 \citep{he2016deep} and transformer \citep{vaswani2017attention}, thus our conclusion is a rather general result.

Finally, we theoretically show that, at a point close to a minimum, the covariance matrix of the noise induced by dropout and the Hessian matrix of the loss landscape is similar in the sense of the expectation with respect to the dropout randomness. The similarity between covariance and Hessian is consistent with experiments, i.e.,  the inverse variance-flatness relation and Hessian-variance alignment relation.

\section{Related works}
Dropout is proposed as a simple way to prevent neural networks from overfitting, and thus improving the generalization of the network \citep{hinton2012improving,srivastava2014dropout}. Many works aim to find an explicit form of dropout. 
\citet{mcallester2013pac} presents PAC-Bayesian bounds, and \citet{wan2013regularization}, \citet{mou2018dropout} derive Rademacher generalization bounds. These results show that the reduction of complexity brought by dropout is $O(p)$, where $p$ is the probability of keeping an element in dropout. \citet{mianjy2020convergence} show that dropout training with logistic loss achieves $\epsilon$-suboptimality in test error in $O(1/\epsilon)$ iterations. All of the above works need specific settings, such as norm assumptions and logistic loss, and they only give a rough estimate of the generalization error bound, which usually consider the worst case. However, it is not clear what is the characteristic of the dropout training process and how to bridge the training with the generalization. In this work, we show that dropout noise has a special structure, which closely relates with the loss landscape. The structure of the effective noise induced by the dropout may be a key reason why dropout can find solutions with better generalization.

Some works attribute the improvement in flatness to the similarity 
between the covariance matrix and the Hessian matrix of the loss function of SGD \citep{papyan2018full,papyan2019measurements}. For example, 
\citet{feng2021inverse} investigate the inverse variance-flatness relation for SGD and \citet{zhu2018anisotropic} study the Hessian-variance alignment for SGD.

\section{Preliminary}

\subsection{Deep Neural Networks}
Consider $L$-layer ($L\geq 2$) fully-connected DNNs with a general differentiable activation function. We regard the input as the $0$th layer and the output as the $L$th layer. Let $m_l$ be the number of neurons in the $l$th layer. In particular, $m_0=d$ and $m_L=d'$. For any $i,k\in \sN$ and $i<k$, we denote $[i:k]=\{i,i+1,\ldots,k\}$. In particular, we denote $[k]:=\{1,2,\ldots,k\}$.
Given weights $W^{[l]}\in \sR^{m_l\times m_{l-1}}$ and bias $b^{[l]}\in\sR^{m_{l}}$ for $l\in[L]$, we define the collection of parameters $\vtheta$ as a $2L$-tuple (an ordered list of $2L$ elements) whose elements are matrices or vectors
\begin{equation*}
    \vtheta=\Big(\vtheta|_1,\cdots,\vtheta|_L\Big)=\Big(\mW^{[1]},\vb^{[1]},\ldots,\mW^{[L]},\vb^{[L]}\Big),
\end{equation*}
where the $l$th layer parameters of $\vtheta$ is the ordered pair $\vtheta|_{l}=\Big(\mW^{[l]},\vb^{[l]}\Big),\quad l\in[L]$.
We may misuse of notation and identify $\vtheta$ with its vectorization $\mathrm{vec}(\vtheta)\in \sR^M$ with $M=\sum_{l=0}^{L-1}(m_l+1) m_{l+1}$.

Given $\vtheta\in \sR^M$, the neural network function $\vf_{\vtheta}(\cdot)$ is defined recursively. First, we write $\vf^{[0]}_{\vtheta}(\vx)=\vx$ for all $\vx\in\sR^d$. Then for $l\in[L-1]$, $\vf^{[l]}_{\vtheta}$ is defined recursively as 
$\vf^{[l]}_{\vtheta}(\vx)=\sigma (\mW^{[l]} \vf^{[l-1]}_{\vtheta}(\vx)+\vb^{[l]})$.
Finally, we denote
\begin{equation*}
    \vf_{\vtheta}(\vx)=\vf(\vx,\vtheta)=\vf^{[L]}_{\vtheta}(\vx)=\mW^{[L]} \vf^{[L-1]}_{\vtheta}(\vx)+\vb^{[L]}.
\end{equation*}
For notational simplicity, we denote

\begin{equation*}
    \vf_{\vtheta}^{j}(\vx_i)=\mW^{[L]}_{j} \vf^{[L-1]}_{\vtheta, j}(\vx_i), 
\end{equation*}
where $\mW^{[L]}_{j} \in  \sR^{m_L}$ is the $j$th column of $\mW^{[L]}$, and $\vf^{[L-1]}_{\vtheta, j}(\vx_i)$ is the $j$th element of vector $\vf^{[L-1]}_{\vtheta}(\vx_i)$.

\subsection{Loss function}
The training data set is denoted as  $S=\{(\vx_i,\vy_i)\}_{i=1}^n$, where $\vx_i\in\sR^d$, $\vy_i\in \sR^{d'}$. For simplicity, here we assume an unknown function $\vy$ satisfying $\vy(\vx_i)=\vy_i$ for $i\in[n]$. The empirical risk reads as
\begin{equation*}
    \RS(\vtheta)=\frac{1}{n}\sum_{i=1}^n\ell(\vf(\vx_i,\vtheta),\vy(\vx_i))=\Exp_S\ell(\vf(\vx,\vtheta),\vy),
\end{equation*}
where the expectation $\Exp_S h(\vx):=\frac{1}{n}\sum_{i=1}^n h(\vx_i)$ for any function $h:\sR^d\to \sR$ and the loss function $\ell(\cdot,\cdot)$ is differentiable and the derivative of $\ell$ with respect to its first argument is denoted by $\nabla\ell(\vy,\vy^*)$. The error with respect to data sample $(\vx_{i},\vy_{i})$ reads as 
\begin{equation*}
    \vepsilon(\vf_{\vtheta}(\vx_{i}),\vy_{i})=\vf_{\vtheta}(\vx_{i})- \vy_{i}.
\end{equation*}
For notational simplicity, we denote $\vepsilon(\vf_{\vtheta}(\vx_{i}),\vy_{i})=\vepsilon_{\vtheta,i}$. 

\subsection{Dropout}
For $\vf_{\vtheta}^{[l]}(\vx) \in \mathbb{R}^{m_{l}}$, we sample a scaling vector $\veta \in \mathbb{R}^{m_{l}}$ with independent random coordinates, 
\begin{equation*}
    (\veta)_{k}= \begin{cases}\frac{1-p}{p} & \text { with probability } p \\ -1 & \text { with probability } 1-p, \end{cases}
\end{equation*}
where $k \in [m_l]$ indexes a coordinate of $\vf_{\vtheta}^{[l]}(\vx)$. Note that $\eta$ is a zero mean random variable. We then apply dropout by computing
\begin{equation*}
\vf_{\vtheta, \veta}^{[l]}(\vx)=(\vone+\veta) \odot \vf_{\vtheta}^{[l]}(\vx),
\end{equation*}
and using $\vf_{\vtheta, \veta}^{[l]}(\vx)$ instead of $\vf_{\vtheta}^{[l]}(\vx)$. Here we use $\odot$ for the Hadamard product of two matrices of the same dimension. With slight abuse of notation, we let $\veta$ denote the collection of such vectors over all layers. $\vf_{\vtheta, \veta}^\mathrm{drop} (\vx)$ denotes the output of model $\vf_{\vtheta} (\vx)$ on input $\vx$ using dropout noise $\veta$. $\RS ^\mathrm{drop}\left(\vtheta, \veta\right)$ denotes the empirical risk with respect to network with dropout layer $\vf_{\vtheta, \veta}^\mathrm{drop}$, i.e.,
\begin{equation*}
    \RS ^\mathrm{drop}\left(\vtheta, \veta\right)= \frac{1}{n}\sum_{i=1}^n\ell(\vf_{\vtheta, \veta}^\mathrm{drop}(\vx_i),\vy(\vx_i))=\Exp_S\ell(\vf_{\vtheta, \veta}^\mathrm{drop}(\vx),\vy).
\end{equation*}

\subsection{Randomness induced by dropout} \label{sec:randomness}
\subsubsection{Random trajectory data}
The training process of neural networks are usually divided into two phases, fast convergence and exploration phase \citep{shwartz2017opening}. 
In this work, we follow the experimental scheme in \citet{feng2021inverse} to show the similarity between dropout and SGD. This can be understood by frequency principle \citep{xu2019training,xu2019frequency,zhang2021linear}, which states that DNNs fast learn low-frequency components but slowly learn high-frequency ones. 

We collect parameter sets $S_{para}=\{\vtheta_{i}\}_{i=1}^{N}$ from $N$ consecutive training steps in the exploration phase, where $\vtheta_{i}$ is the network parameter set at  $i$th sample point. 
\subsubsection{Random gradient data}
We often need larger time interval for enough sampling to estimate the covariance accurately. Although the network loss is small, compared with the initial sampling parameters, the network parameters could have large changes during the long-time sampling. Therefore, much extra noise may be induced. Meanwhile, for dropout, it is difficult to get a small loss value on large networks and datasets, therefore, model parameters often have large fluctuations during the sampling. To overcome this problem, we propose a more appropriate sampling method to avoid additional noise caused by sampling parameters in a large time interval.
We train the network until the loss is small enough and then freeze the training. We sample $N$ gradients of the loss function w.r.t. the parameters with different dropout variables, i.e., $S_{grad}=\{\vg_{i}\}_{i=1}^{N}$. In each sample, the dropout rate is fixed. In this way, we can get the noise structure of dropout without being affected by parameter changes caused by long-term training.

\subsection{Inverse variance-flatness relation}
We study the inverse variance flatness relation for both random trajectory data and random gradient data. For convenience, we denote data as $S$ and its covariance as $\Sigma$.
\subsubsection{Variance vs. interval flatness}
The definitions of variance and interval flatness are as follows:
\begin{definition}[\textbf{Variance of data at an eigen direction}]
    For data $S$ and its covariance $\Sigma$, by denoting $\lambda_{i}(\Sigma)$ as the $i$th eigenvalue of $\Sigma$, we write $\lambda_{i}(\Sigma)$ as the variance of the data at the corresponding eigen direction.
\end{definition}

\begin{definition}[\textbf{Interval flatness}]\footnote{This definition is also used in \citet{feng2021inverse}}
   For a specific solution $\vtheta^{*}_{0}$, the loss function profile $L_{\vv}$ along the direction $\vv$ is:
$$L_{\vv}(\delta \theta)\equiv L(\vtheta^{*}_{0}+\delta \theta \vv), $$ 
 where $\delta \theta$ represents the distance moved in the $\vv$ direction. The interval flatness $F_{\vv}$ is defined as the width of the region within which $L_{\vv}(\delta \theta)\leq 2L_{\vv}(0)$. We determine $F_{\vv}$ by finding two closest points $\theta_{\vv}^{l}<0$ and $\theta_{\vv}^{r}>0$ on each side of the minimum that satisfy $L_{\vv}(\theta_{\vv}^{l})=L_{\vv}(\theta_{\vv}^{r})=2L_{\vv}(0)$. The interval flatness is defined as:
\begin{equation}
  F_{\vv}\equiv \theta_{\vv}^{r}-\theta_{\vv}^{l}.
\end{equation}

\end{definition}

\begin{remark}
    The experiments show that the result is not sensitive to the selection of the pre-factor 2. A larger value of $F_{\vv}$ means a flatter landscape in the direction $\vv$.
\end{remark}

Denote $\lambda_{i}(\Sigma)$ as the $i$th eigenvalue of $\Sigma$, and denote its corresponding eigen-vector as $\vv_{i}(\Sigma)$. The interval flatness of the loss landscape in the direction $\vv_{i}(\Sigma)$ is denoted as $F_{\vv_{i}(\Sigma)}$. We then experimentally explore the relation of $\{\lambda_{i}(\Sigma), F_{\vv_{i}(\Sigma)}\}_{i=1}^{N}$. 
\subsubsection{Projected variance vs. Hessian flatness}

The definitions of projected variance and Hessian flatness are as follows:

\begin{definition}[\textbf{Projected variance}]
    For a given direction $\bm{v}\in \sR^{d_{\vtheta}}$ and a parameter set $S=\{\vtheta_{i}\}_{i=1}^{n_{S}}$, where $\vtheta_{i} \in \sR^{d_{\vtheta}}$, the inner product of $\bm{v}$ and $\vtheta_{i}$ is denoted by $ {\rm Proj}_{\bm{v}}(\vtheta_{i}):=\vtheta^{T}_{i}\bm{v}$, then we can define the projected variance at direction $\bm{v}$ with respect to the sample set $S$ as follows,
    \begin{equation*}
        {\rm Var}({\rm Proj}_{\bm{v}}(S))=\frac{\sum_{i=1}^{n_{S}}({\rm Proj}_{\bm{v}}(\vtheta_{i})-\mu)^2}{n_{S}}, 
    \end{equation*}
    where $\mu$ is the mean value of $\{{\rm Proj}_{\bm{v}}(\vtheta_{i})\}_{i=1}^{n_{S}}$.
\end{definition}

\begin{definition}[\textbf{Hessian flatness}]
    For Hessian matrix $H$, by denoting $\lambda_{i}(H)$ as the $i$th eigenvalue of $H$, we write $\lambda_{i}(H)$ as the Hessian flatness.
\end{definition}

To obtain the variance induced by the dropout at a fixed position $\vtheta$, we propose another way to characterize the inverse variance-flatness relation. For given data $S$ and Hessian matrix $H$, we experimentally explore the relation of $\{{\rm Var}({\rm Proj}_{\bm{v_i(H)}}(S)), \lambda_{i}(H)\}_{i=1}^{N}$, where $\lambda_{i}(H)$ and $\vv_i(H)$ is the $i$th eigenvalue and eigenvector of $H$, respectively.

\subsection{Hessian-variance alignment}

Similar to \citet{zhu2018anisotropic}, we quantify the alignment between the noise structure and the curvature of loss landscape by 
$$T_i = \operatorname{Tr}(H_i \Sigma_i), $$
where $\Sigma_i$ is the $i$th-step covariance matrix of dropout layers and $H_i$ is the Hessian matrix of the loss landscape at network parameters of the $i$th-step. 

\begin{table}[t]
\caption{Three types of experiments explain why dropout finds flat minima. }
\label{sample-table}
\centering
\vskip 0.15in
\begin{center}
\begin{small}
\begin{sc}
\begin{tabular}{m{2cm}<{\centering}m{2.5cm}<{\centering}m{2.5cm}<{\centering}}
\toprule
&\multicolumn{2}{c}{Dropout covariance $\Sigma$} \\
\cmidrule(lr){2-3}
 & Trajectory variance $\Sigma_t$ & gradient variance $\Sigma_g$  \\
\midrule
Interval flatness $F_{\vv}$    & \multicolumn{2}{c}{$\{\lambda_{i}(\Sigma), F_{\vv_{i}(\Sigma)}\}$, Fig. \ref{fig:pca}} \\
\midrule
\multirow{2}{*}{\tabincell{c<{\centering}}{ \\Hessian \\ flatness $\lambda(H)$ }} & \multicolumn{2}{c}{$\{{\rm Var}({\rm Proj}_{\vv_i(H)}(S)),\lambda_{i}(H)\}$, Fig. \ref{fig:Hessian}}\\ 
\cmidrule(lr){2-3}
    &  $\backslash$ & Alignment: $Tr(H\Sigma_g)$, Fig. \ref{fig:anisotropic}.\\

\bottomrule
\end{tabular}\label{tab:cifar10}
\end{sc}
\end{small}
\end{center}
\vskip -0.1in
\end{table}

\section{Experimental setup} \label{sec:setup}

To understand the effect of dropout, we train a number of networks with different structures. We consider the following types of neural networks: 1) Fully-connected neural networks (FNNs) trained by MNIST \citep{lecun1998gradient}. 2) Convolutional neural networks (CNNs) trained by CIFAR-10 \citep{krizhevsky2009learning}. 3) Deep residual neural networks (ResNets) \citep{he2016deep} trained by CIFAR-100 \citep{krizhevsky2009learning}. 4) Transformer \citep{vaswani2017attention} trained by Multi30k \citep{elliott2016multi30k}. The loss of all our experiments is cross entropy loss.

It is worth noting that, to avoid the influence of SGD in our experiments, all our networks are trained using GD, so it is difficult for us to verify on larger datasets such as ImageNet.

The detailed experimental setup can be found in Appendix \ref{appendix:expsetup}.

\section{Dropout finds flatter minima}
Dropout is almost ubiquitous in training deep networks. It is interesting and important to understand what makes dropout improve the generalization of training neural networks. Inspired by the study of SGD \citep{keskar2016large}, we explore the flatness of the minima found by dropout. 


We adopt the method of \citet{li2017visualizing} in this work as follows. 
To obtain a direction for a network with parameters $\vtheta$, we begin by producing a random Gaussian direction vector $\vd$ with dimensions compatible with $\vtheta$. Then, we normalize each filter in $\vd$ to have the same norm of the corresponding filter in $\vtheta$. For FNN, each layer can be regarded as a filter. The normalization process is equivalent to normalizing the layer. For CNN, each convolution kernel may have multiple filters. Each filter is normalized individually. In other words, we make the replacement $\vd_{i, j} \leftarrow \frac{\vd_{i, j}}{\left\|\vd_{i, j}\right\|}\left\|\vtheta_{i, j}\right\|$, where $\vd_{i,j}, \vtheta_{i, j}$ represent the $j$th filter of the $i$th layer of the random direction $\vd$ and the network parameters $\vtheta$, and $\|\cdot\|$ denotes the Frobenius norm. It should be noted that $j$ is not the index of the weight, but the filter. We use $f(\alpha)=L\left(\vtheta+\alpha \vd\right)$ to characterize the loss landscape around the minima obtained with dropout layers $\vtheta^*_{\vd}$ and without dropout layer $\vtheta^*$. 

\begin{figure}[h]
	\centering
	\subfigure[flatness of FNN]{\includegraphics[width=0.24\textwidth]{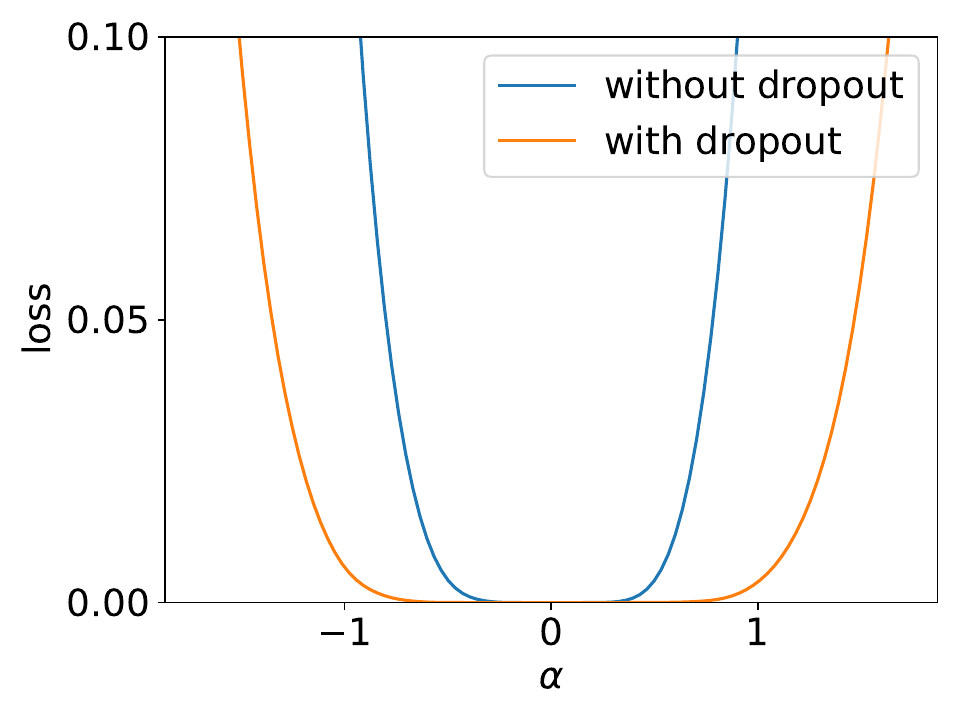}}
	\subfigure[flatness of vgg-9]{\includegraphics[width=0.24\textwidth]{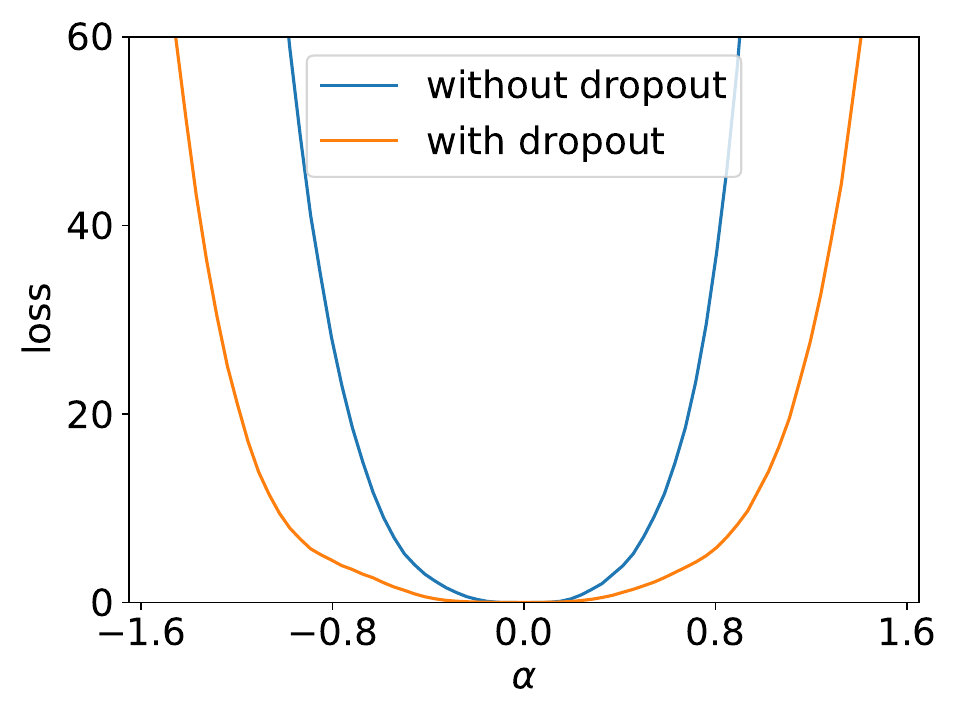}}
	\subfigure[flatness of ResNet-20]{\includegraphics[width=0.24\textwidth]{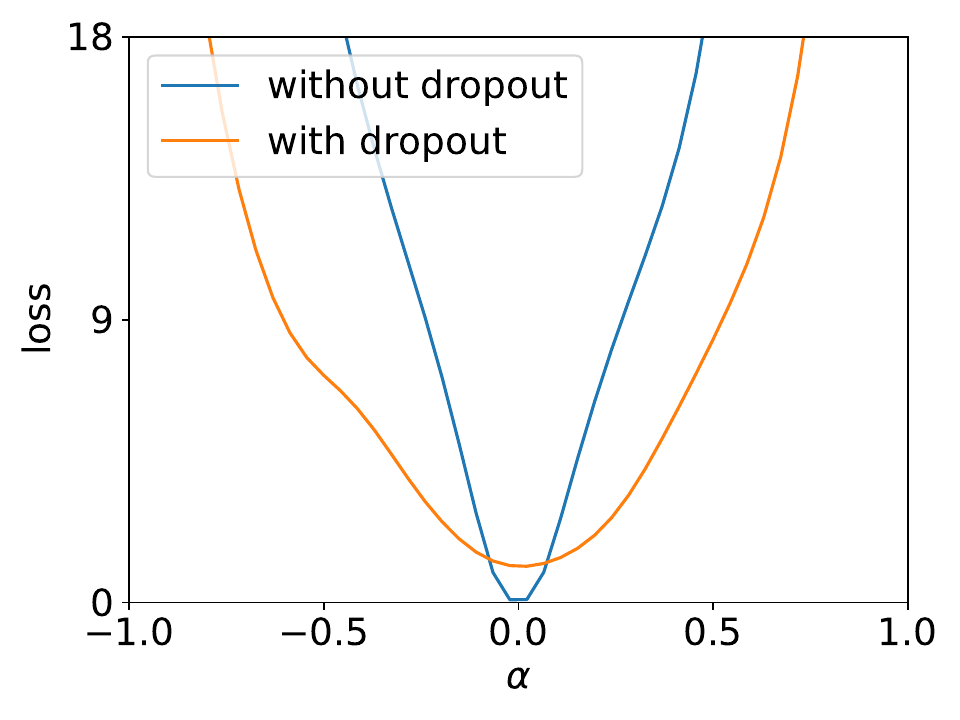}}
	\subfigure[flatness of transformer]{\includegraphics[width=0.24\textwidth]{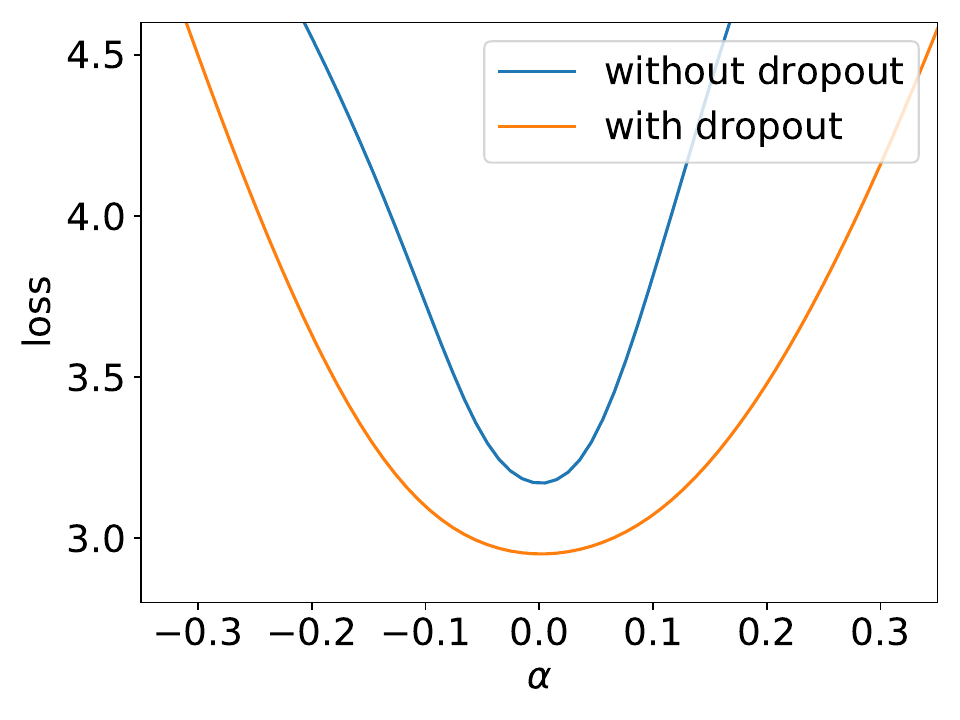}}
  \caption{The 1D visualization of solutions of different network structures obtained with or without dropout layers. (a) The FNN is trained on MNIST dataset. The test accuracy for model with dropout layers is $98.7\%$ while $98.1\%$ for model without dropout layers. (b) The vgg-9 network is trained on CIFAR-10 dataset using the first 2048 examples as training dataset. The test accuracy for model with dropout layers is $60.6\%$ while $59.2\%$ for model without dropout layers. (c) The ResNet-20 network is trained on CIFAR-100 dataset using all examples as training dataset. The test accuracy for model with dropout layers is $54.7\%$ while $34.1\%$ for model without dropout layers. (d) The transformer is trained on the Multi30k dataset using the first 2048 examples as training dataset. The test accuracy for model with dropout layers is $49.33\%$ while $34.73\%$ for model without dropout layers.}\label{fig:flatness_cnn}
\end{figure}

For all network structures shown in Fig. \ref{fig:flatness_cnn}, dropout can improve the generalization of the network and find a flatter minimum. In Fig. \ref{fig:flatness_cnn}(a, b), for both networks trained with and without dropout layers, the training loss values are all closed to zero, but their flatness and generalization are still different. In Fig. \ref{fig:flatness_cnn}(c, d), due to the complexity of the dataset, i.e., CIFAR-100 and Multi30k, and networks, i.e., ResNet-20 and transformer, networks with dropout layers does not achieve very small training error but the ones with dropout find flatter minima with much better generalization.

Next, we utilize three methods to examine the relation between the covariance of the noise induced by the dropout randomness and the Hessian of the loss landscape, as summarized in Table \ref{tab:cifar10}.

\section{Inverse variance-flatness relation}
Similar to SGD, the effect of dropout can be equivalent to imposing a specific noise on the gradient. A random noise, such as isotropic noise, can help the training escape local minima, but can not robustly improve generalization \citep{an1996effects,zhu2018anisotropic}. The noise induced by the dropout should have certain properties that can lead the training to good minima. 

In this section, we show that the noise induced by the dropout satisfies the inverse variance-flatness relation, that is, the noise variance is larger along the sharper direction of the loss landscape at a minimum. The landscape-dependent structure helps the training escape sharp minima. 


\subsection{Variance vs. interval flatness}
We use the principal component analysis (PCA) to study the weight variations when the accuracy is nearly $100\%$. 
For FNNs, networks are trained on MNIST with the first 10000 examples as the training set for computational efficiency. For ResNets, networks are trained on CIFAR-100 with 50000 examples as the training set. For the transformer structure, the network is trained by Multi30k \citep{vaswani2017attention}.
The networks are trained with full batch for different learning rates and dropout rates under the same random seed (that is, with the same initialization parameters). When the loss is small enough, we sample the parameters or gradients of parameters $N$ times ($N=3000$ in this experiment) and use the method introduced in Section \ref{sec:randomness} to construct covariance matrix $\Sigma$ by the
weights $S_{para}$ or gradients $S_{grad}$ of specific network parameters mentioned in Section \ref{sec:setup}. The PCA is done for the covariance matrix $\Sigma$. We then compute the interval flatness of the loss function landscape at eigen-directions, i.e., $\{F_{\vv_{i}(\Sigma)}\}_{i=1}^N$. Note that the PCA spectrum $\{\lambda_{i}(\Sigma)\}_{i=1}^N$ indicate the variance of weights $S_{para}$ or gradients $S_{grad}$ at corresponding eigen-directions. 

\begin{figure}[h]
	\centering
	\subfigure[FNN, datasets $S$ sampled from parameters]{\includegraphics[width=0.3\textwidth]{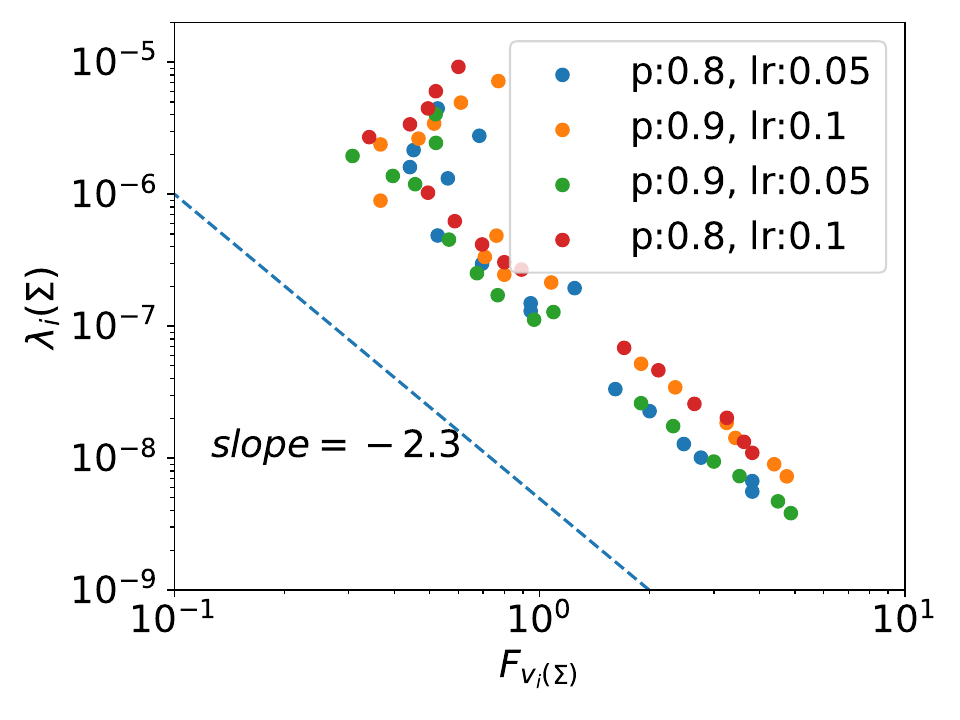}}
	\subfigure[ResNet-20, datasets $S$ sampled from parameters]{\includegraphics[width=0.3\textwidth]{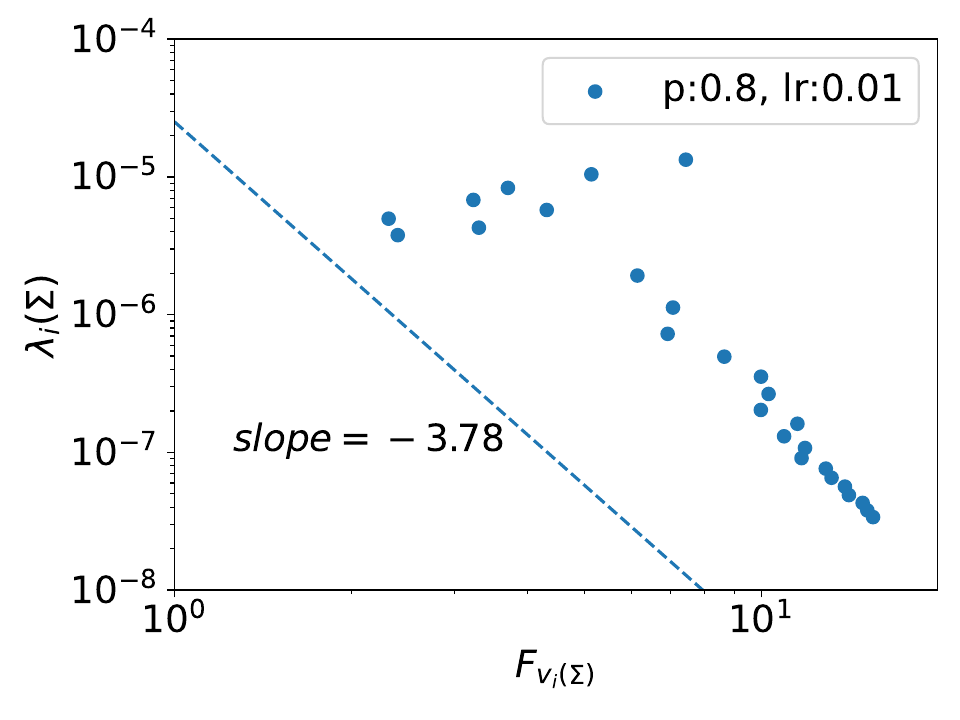}}
	\subfigure[Transformer, datasets $S$ sampled from parameters]{\includegraphics[width=0.3\textwidth]{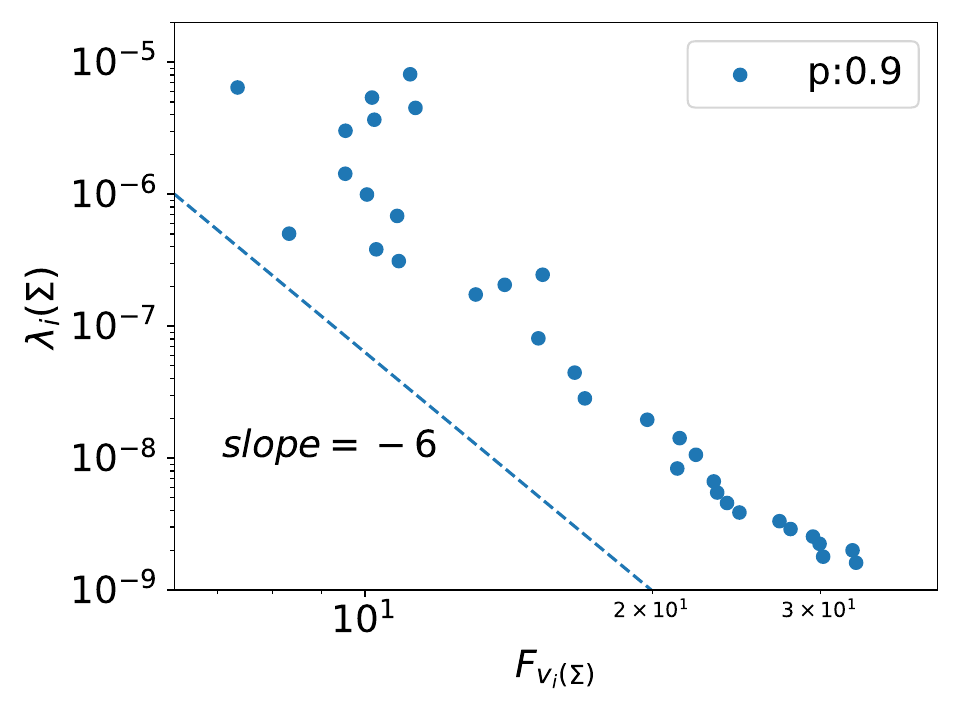}}
	
	\subfigure[FNN, datasets $S$ sampled from gradients of parameters]{\includegraphics[width=0.3\textwidth]{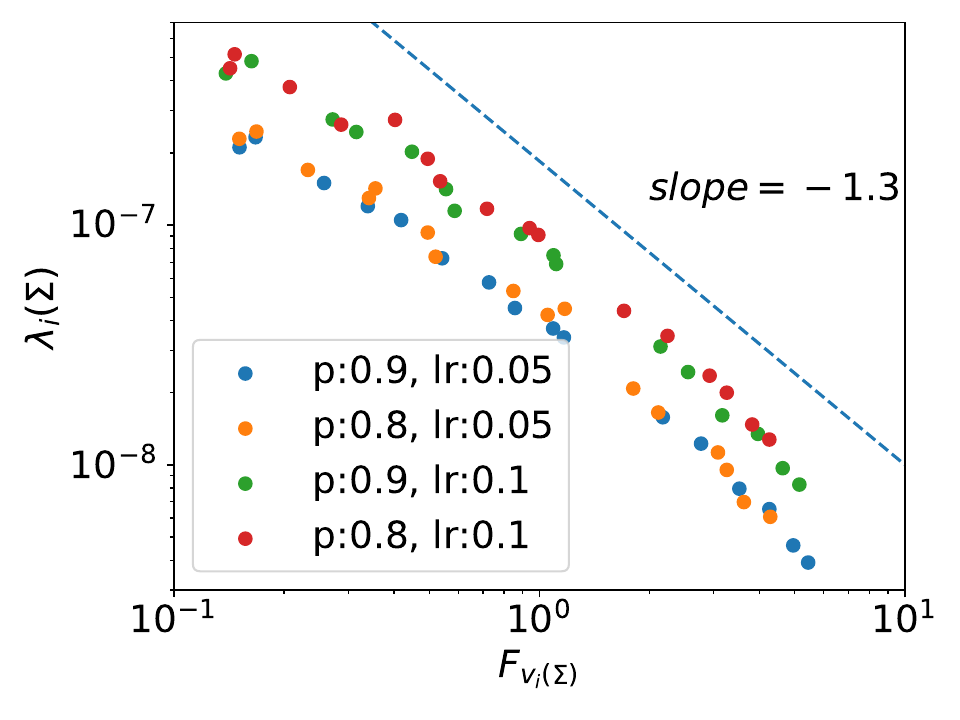}} 
	\subfigure[ResNet-20, datasets $S$ sampled from gradients of parameters]{\includegraphics[width=0.3\textwidth]{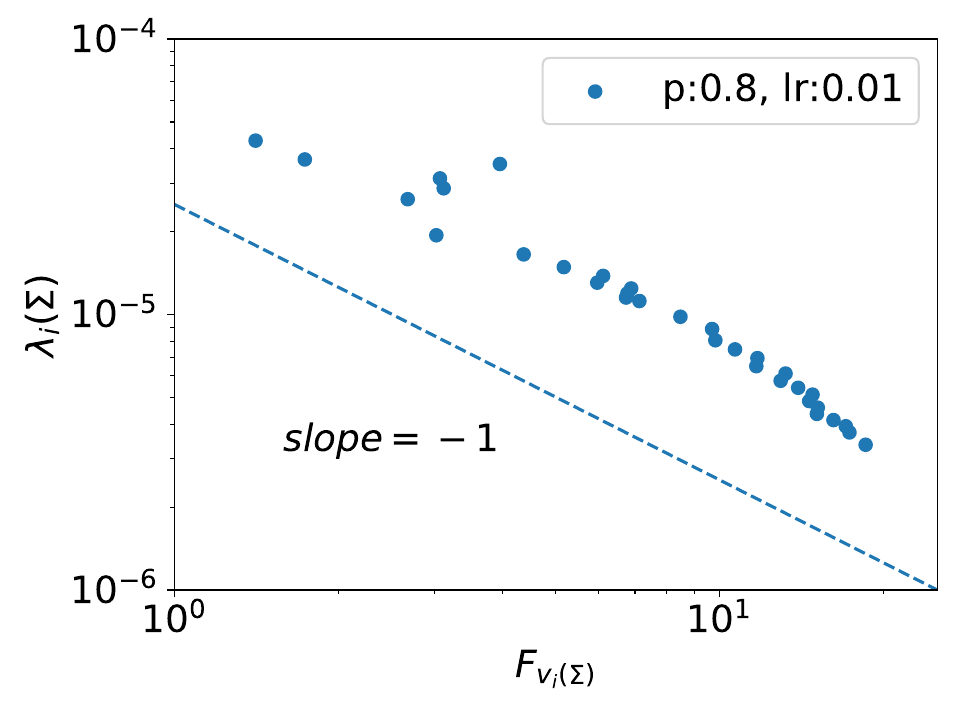}} 
	\subfigure[Transformer, datasets $S$ sampled from gradients of parameters]{\includegraphics[width=0.3\textwidth]{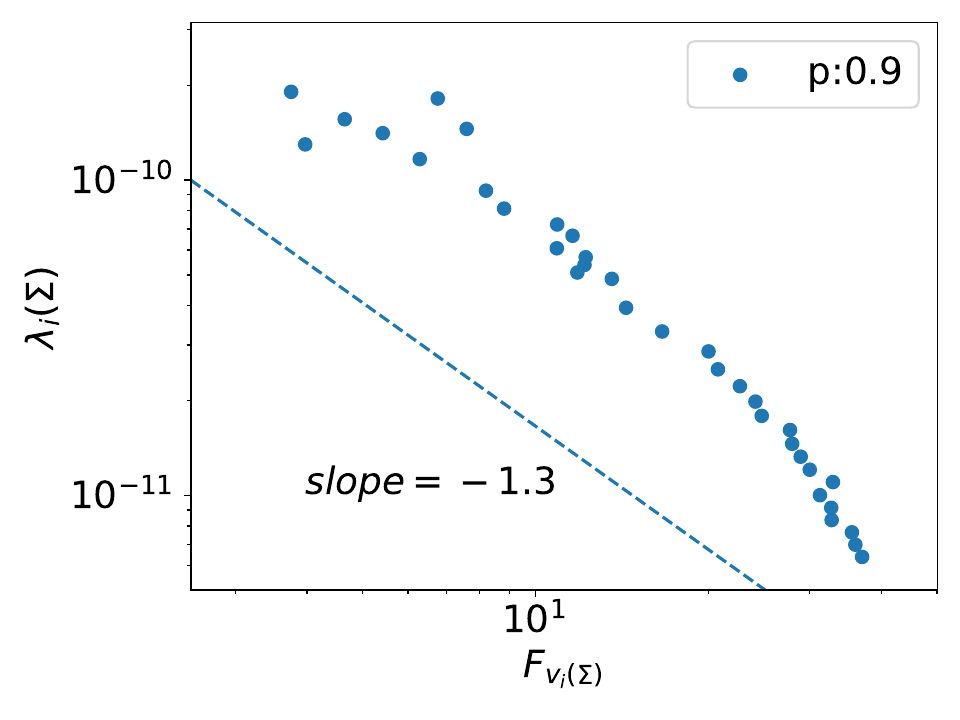}} 
  \caption{The inverse relation between the variance $\{\lambda_{i}(\Sigma)\}_{i=1}^N$ and the interval flatness $\{F_{\vv_{i}(\Sigma)}\}_{i=1}^N$ for different choices of dropout rate $p$ and learning rate $lr$ with different network structures. The PCA is done for different datasets $S$ sampled from parameters for the top line and sampled from gradients of parameters for the bottom line. The dash lines give the approximate slope of the scatter.} \label{fig:pca}
\end{figure} 

As shown in Fig. \ref{fig:pca}, for different learning rates and dropout rates, there is an inverse relationship between the interval flatness of the loss function landscape $\{F_{\vv_{i}(\Sigma)}\}_{i=1}^N$ and the dropout variance, i.e., the PCA spectrum $\{\lambda_{i}(\Sigma)\}_{i=1}^N$. We can approximately see a power-law relationship between $\{F_{\vv_{i}(\Sigma)}\}_{i=1}^N$ and $\{\lambda_{i}(\Sigma)\}_{i=1}^N$. More detailed, for the small flatness part, the variance of noise induced by dropout is generally large, which indicates that the noise induced by dropout has larger variance in sharp directions; for the large flatness part, as the loss landscape gets flatter, the linear relationship is more obvious, and we can see a clearer asymptotic behavior in the results. Overall, we can observe the negative correlation between the 
variance and flatness in Fig. \ref{fig:pca}.



\subsection{Projected variance vs. Hessian flatness}

The eigenvalues of the Hessian of the loss at a minimum are also often used to indicate the flatness. A large eigenvalue corresponds to a sharper direction. In this section, we study the relationship between eigenvalues of Hessian $H$ of loss landscape at the end point of training and the variances of dropout at corresponding eigen-directions. 
As mentioned in the Preliminary, we sample the parameters or gradients of parameters 1000 times, that is $N=1000$. For each eigen-direction $\vv_i$ of Hessian $H$, we project the sampled parameters or the gradients of sampled parameter to direction $\vv_i$ by inner product, denoted by ${\rm Proj}_{\vv_i}(S)$. Then, we compute the variance of the projected data, i.e., ${\rm Var}({\rm Proj}_{\vv_i}(S))$. 

\begin{figure}[h]
	\centering
	\subfigure[FNN, FNN, datasets $S$ sampled from parameters]{\includegraphics[width=0.3\textwidth]{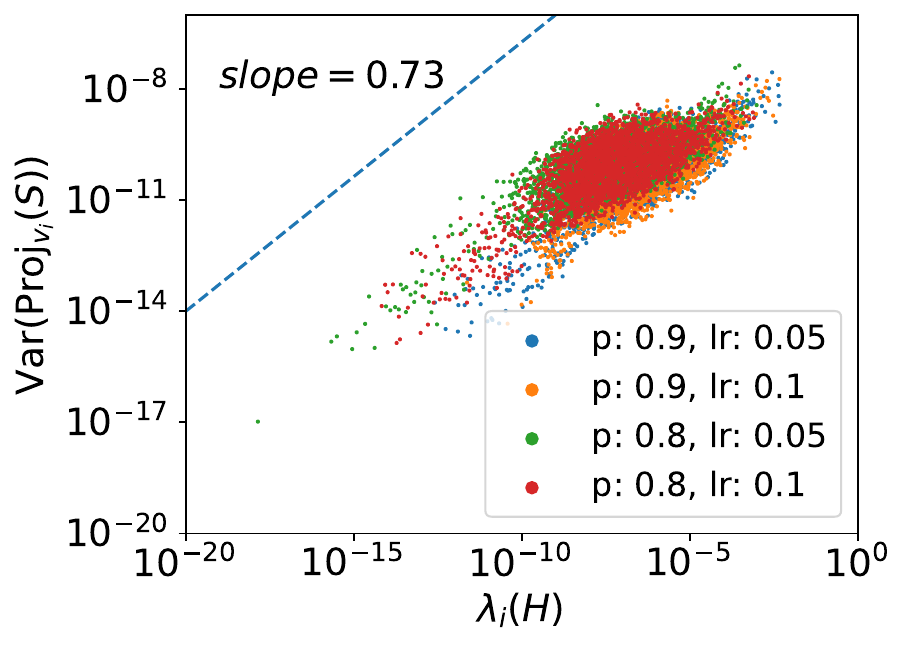}}
	\subfigure[ResNet-20, datasets $S$ sampled from parameters]{\includegraphics[width=0.3\textwidth]{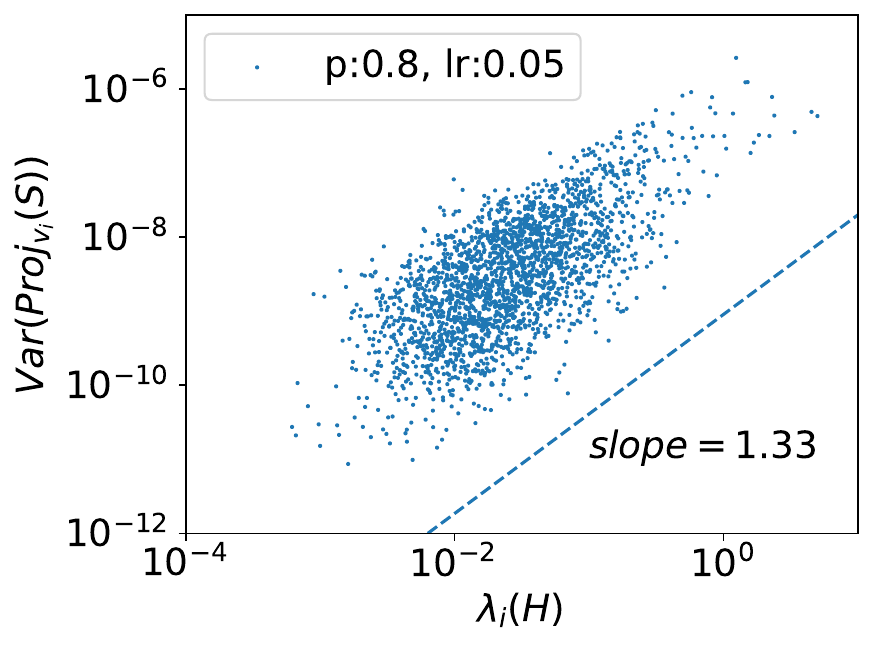}}
	\subfigure[Transformer, datasets $S$ sampled from parameters]{\includegraphics[width=0.3\textwidth]{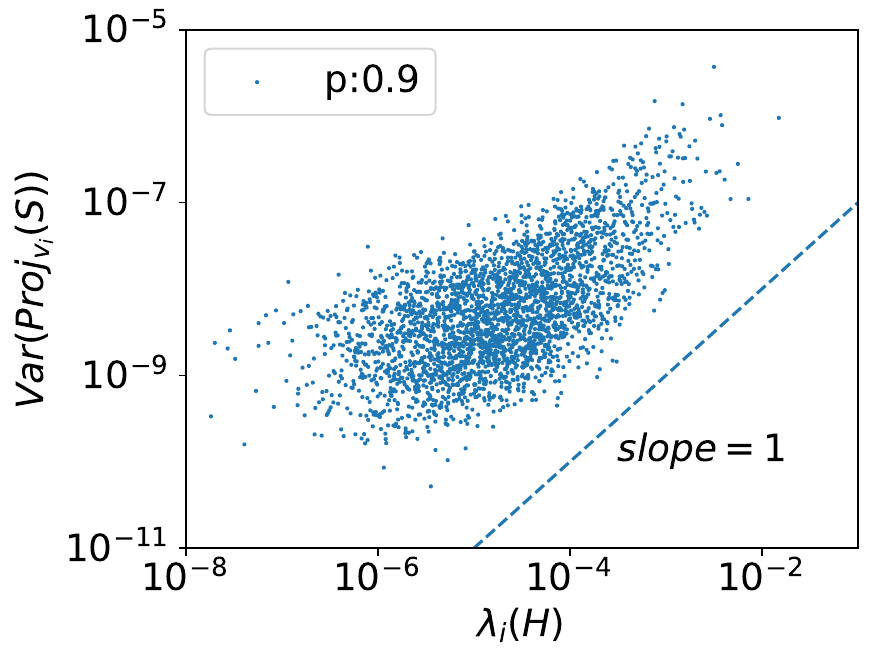}}
	
	\subfigure[FNN, datasets $S$ sampled from gradients of parameters]{\includegraphics[width=0.3\textwidth]{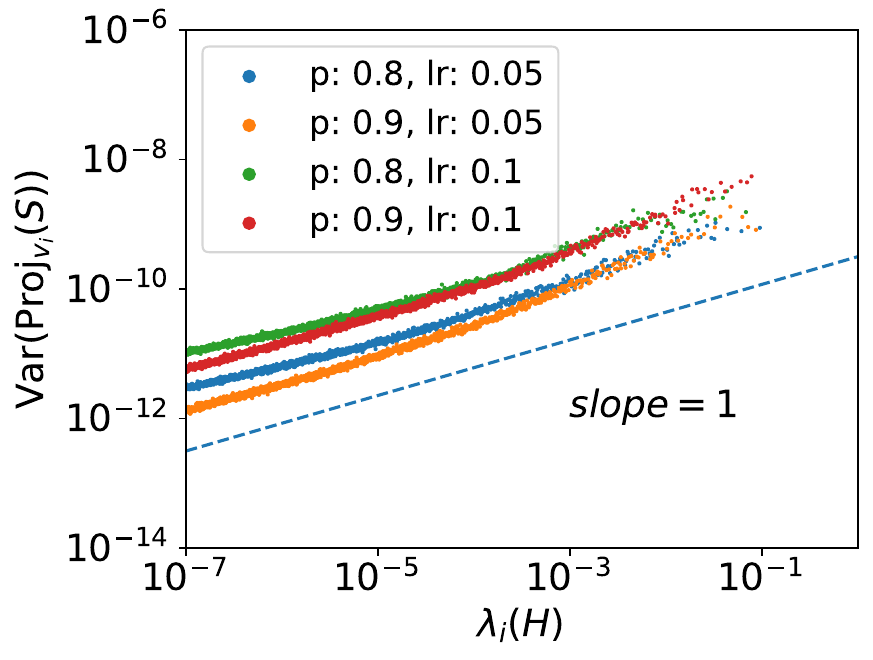}} 
	\subfigure[ResNet-20, datasets $S$ sampled from gradients of parameters]{\includegraphics[width=0.3\textwidth]{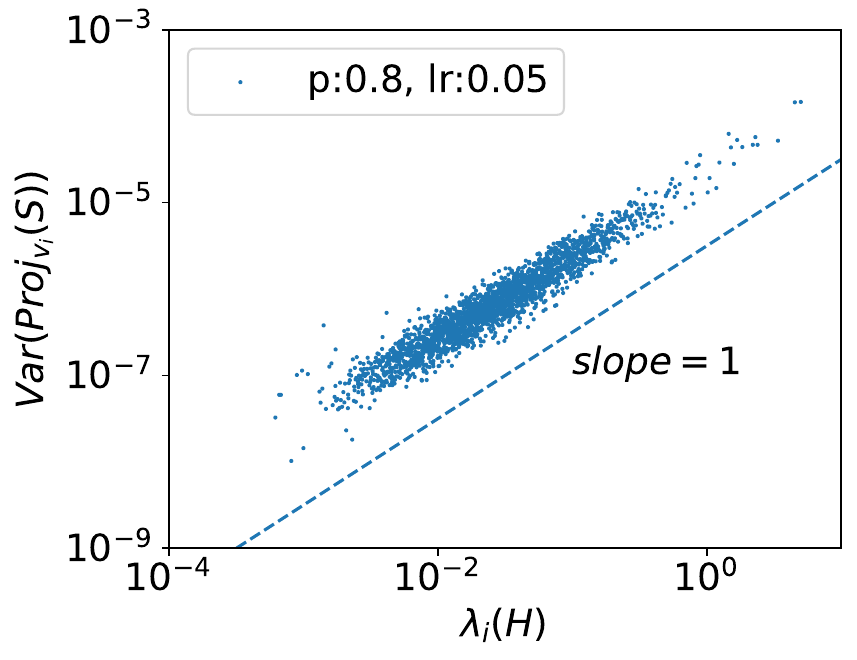}}
	\subfigure[Transformer, datasets $S$ sampled from gradients of parameters]{\includegraphics[width=0.3\textwidth]{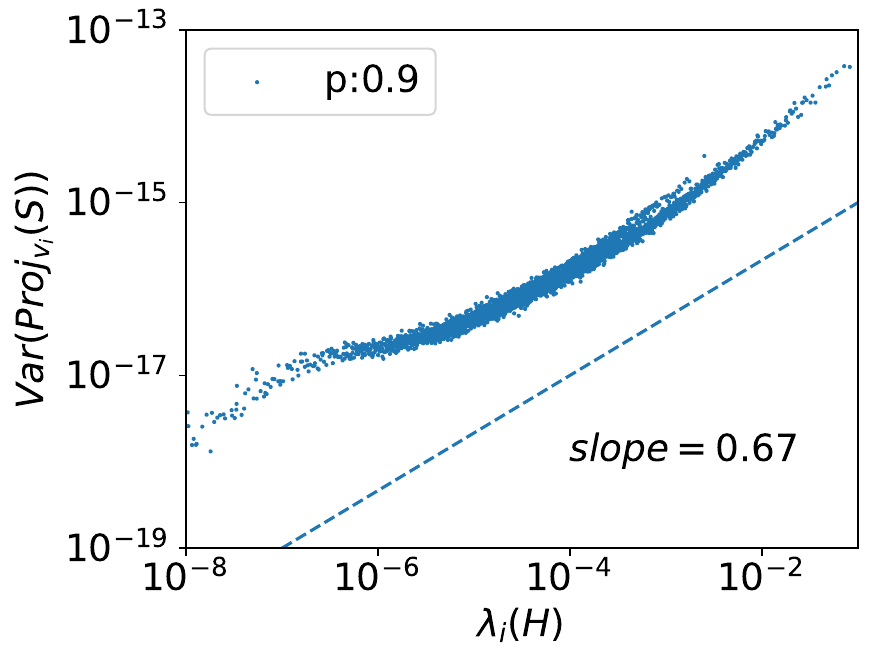}} 
  \caption{The relation between the variance $\{{\rm Var}({\rm Proj}_{\vv_i}(S))\}_{i=1}^D$ and the eigenvalue $\{\lambda_{i}(H)\}_{i=1}^{D}$ for different choices of dropout rate $p$ and learning rate $lr$ with different network structures. The projection is done for different datasets $S$ sampled from parameters for the top line and sampled from gradients of parameters for the bottom line. The dash lines give the approximate slope of the scatter.}\label{fig:Hessian}
\end{figure}

As shown in Fig. \ref{fig:Hessian}, we find that there is also a power-law relationship between $\{\lambda_{i}(H)\}_{i=1}^{D}$ and $\{{\rm Var}({\rm Proj}_{\vv_i}(S))\}_{i=1}^D$ for different dropout rates and learning rates, no matter $S$ is sampled from parameters or gradients of parameters. The positive correlation between the eigenvalue and the projection variance show the structure of the dropout noise, which helps the network escape the bad minima. At the same time, as shown in Figs. \ref{fig:pca} and \ref{fig:Hessian}, we can see that gradient sampling has a more clear linear structure than that of parameter sampling.

\section{Hessian-variance alignment}
In this section, we study the alignment between the Hessian and the random gradient covariance at each training step, i.e., Hessian-variance alignment. 
Note that the training is performed by GD without dropout. At step $i$, we sample the gradients of parameters $\{\vg_i^j\}_{j=1}^{N}$ by tentatively adding a dropout layer between the hidden layers. For each step $i$, we the compute $\operatorname{Tr}(H_i \Sigma_i)$, where $H_i$ is the Hessian of the loss at the parameter set at step $i$ and $\Sigma_i$ is the covariance of $\{\vg_i^j\}_{j=1}^{N}$.

\begin{figure}[h]
	\centering
	\includegraphics[width=0.5\textwidth]{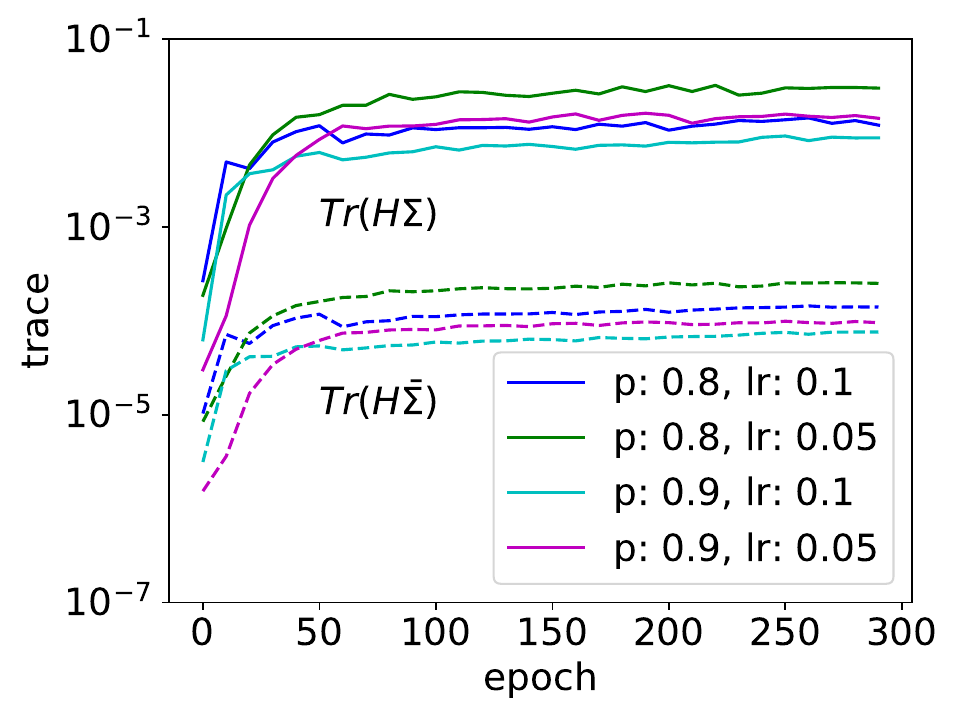}
	\caption{Comparison between $\operatorname{Tr}(H_i \Sigma_i)$ and $\operatorname{Tr}(H_i \bar{\Sigma_i})$ in each training epoch $i$ for different choices of dropout rate $p$ and learning rate $lr$. The FNN is trained on MNIST dataset using the first 10000 examples as training dataset. The solid and the dotted lines represent the value of $\operatorname{Tr}(H_i \Sigma_i)$ and $\operatorname{Tr}(H_i \bar{\Sigma_i})$, respectively. \label{fig:anisotropic}}
\end{figure}

In order to show the anisotropic structure, we construct the isotropic noise for comparison, i.e., $\bar{\Sigma_i}=\frac{\operatorname{Tr} \Sigma_i}{D} I$ of the covariance matrix $\Sigma_i$, where $D$ is the number of parameters. In our experiments, $D=2500$. As shown in Fig. \ref{fig:anisotropic}, in the whole training process under different learning rates and dropout rates, $\operatorname{Tr}(H_i \Sigma_i)$ is much larger than $\operatorname{Tr}(H_i \bar{\Sigma_i})$, indicating the anisotropic structure of dropout noise and its high alignment with the Hessian matrix. 

\section{Theoretical analysis}\label{sec:theo}
In this section, we summarize key theoretical results on the similarity between Hessian and covariance matrices under dropout regularization. The proofs are in the Appendix \ref{app:1}. We first summarize the specific settings and the assumptions required for our theoretical results:

\begin{setting}[\textbf{Dropout structure}]\label{set:1}
    Consider a $L$-layer ($L\geq 2$) fully-connected DNN has only one dropout layer after the $L-1$th layer of the network,  
    \begin{equation*}
        \vf_{\vtheta, \veta}^{drop} (\vx)=\mW^{[L]}(\vone+\veta) \odot \vf^{[L-1]}_{\vtheta}(\vx)+\vb^{[L]}.
    \end{equation*}
\end{setting}

\begin{setting}[\textbf{Loss function}]\label{set:2}
    Take the mean squared error (MSE) as our loss function, 
    \begin{equation*}
        \RS(\vtheta)=\Exp_S\ell(\vf(\vx,\vtheta),\vy)=\frac{1}{2n}\sum_{i=1}^n(\vf(\vx_i,\vtheta)-\vy_{i})^2.
    \end{equation*}
\end{setting}

\begin{setting}[\textbf{Network structure}]\label{set:3}
    Take the piece-wise linear function as our activation function. For convenience, we further set that the model output is an one-dimensional vector, i.e. $m_{L}=1$.
\end{setting}

\begin{assumption}\label{ass:1}
    We examine the loss landscape after training reaches a stable stage, so we assume that the gradient of the loss function in the average sense is small enough, i.e.
    \begin{equation*}
        \nabla_{\vtheta} \Exp_{\veta}R_{S}^{\mathrm{drop}}(\vtheta,\veta)\approx \mzero.
    \end{equation*}
\end{assumption}

Under the above assumptions and settings, we can theoretically calculate the Hessian matrix and covariance matrix of the loss function as follows.

\begin{thm}[\textbf{Hessian matrix with dropout regularization}]Based on the Setting \ref{set:1}-\ref{set:3} and Assumption \ref{ass:1}, the Hessian matrix of the loss function with respect to $\vf_{\vtheta,\veta}^{\mathrm{drop}}(\vx)$ can be written in the mean sense as:
\begin{equation*}
    H(\vtheta)=\Exp_{\veta}\nabla_{\vtheta}^{2}R_{S}^{\mathrm{drop}}=\frac{1}{n} \sum_{i=1}^{n}\left[ \nabla_{\vtheta} \vf_{\vtheta}\left(\vx_{i}\right) \nabla_{\vtheta}^{\T} \vf_{\vtheta}\left(\vx_{i}\right)+ \frac{1-p}{p} \sum_{j=1}^{m_{L-1}} \nabla_{\vtheta} \vf_{\vtheta}^{j}\left(\vx_{i}\right)  \nabla_{\vtheta}^{\T} \vf_{\vtheta}^{j}\left(\vx_{i}\right) \right].
\end{equation*}


\label{thm:1}
\end{thm}

\begin{thm}[\textbf{Covariance matrix with dropout regularization}]Based on the Setting \ref{set:1}-\ref{set:3} and Assumption \ref{ass:1}, the covariance matrix of the loss function under the randomness of dropout variable $\veta$ and data $\vx$ can be written as:
\begin{equation*}
    \begin{aligned}
    \Sigma^{\mathrm{drop}}_{\vtheta}&=\frac{1}{n}\sum_{i=1}^{n} \left[ a_i \nabla_{\vtheta}\vf_{\vtheta}(\vx_i)\nabla_{\vtheta}^{\T}\vf_{\vtheta}(\vx_i)+ b_{i} \frac{1-p}{p}\sum_{j=1}^{m_{L-1}} \nabla_{\vtheta}\vf_{\vtheta}^{j}(\vx_i)\nabla_{\vtheta}^{\T}\vf_{\vtheta}^{j}(\vx_i)\right],
    \end{aligned}
\end{equation*}
where $a_i=2\Exp_{\veta}\ell(\vf_{\vtheta,\veta}^{\mathrm{drop}}(\vx_i),\vy_{i})$, $b_{i}=2\ell(\vf_{\vtheta}(\vx_i),\vy_{i})$.

\label{thm:2}
\end{thm}

\begin{prop}\label{prop:1}
Based on the Setting \ref{set:1}-\ref{set:3} and Assumption \ref{ass:1}, we further restrict the problem to a binary classification problem, i.e. $\vy_{i} \in \{0,1\},\ \forall i \in [n]$, and assume the model output $\vf_{\vtheta}(\vx_i) \in [\delta, 1-\delta] $ (we can limit the network output using a threshold activation function), where $\delta$ is a small positive constant, then we have:

(i) $\Sigma^{\mathrm{drop}}_{\vtheta} \succeq \delta^{2} H(\vtheta)$, almost everywhere in $\sR^{M}$, $M$ is the dimension of $\vtheta$; 

(ii) For any $\epsilon>0$, and a network parameter $\vtheta\in\Omega=\{\vtheta: \Exp_{\veta}\ell(\vf_{\vtheta,\veta}^{\mathrm{drop}}(\vx_i),\vy_{i}) \leq  \frac{ (\delta+\epsilon)^{2}}{2},\ell(\vf_{\vtheta}(\vx_i),\vy_{i}) \leq  \frac{ (\delta+\epsilon)^{2}}{2},  \forall i \in [n]\}$, we have $\Sigma^{\mathrm{drop}}_{\vtheta} \preceq  (\delta+\epsilon)^{2}  H(\vtheta)$ almost everywhere in $\Omega$. 
\end{prop}
\begin{remark}
$A \preceq B$ means that $(B-A)$ is semi-positive definite.
\end{remark}
\begin{remark}
    The results in Proposition \ref{prop:1} are consist with the results under SGD setting studied in  \citet{papyan2018full,papyan2019measurements,zhu2018anisotropic}.
\end{remark}
From above analysis, we can see the Hessian and the covariance are very similar. Especially, when the training is approaching the end, the error of all samples has similar magnitude, then, $H$ and $\Sigma^{\mathrm{drop}}_{\vtheta}$ has an approximately linear relation.

\section{Conclusion and discussion} \label{sec:dis}

In this work, we find that dropout training selects flatter minima compared with standard gradient descent training. We further show inverse variance-flatness relation and Hessian-variance alignment. These two relations may help the training select flatter minima and leads the training to good generalization. We then theoretically show the similarity between the Hessian and covariance to further support the goodness of dropout. The dropout and the SGD are common in sharing the these two relations. As a starting point, our work shows a promising and reasonable direction for understanding the stochastic training of neural networks.




\bibliographystyle{elsarticle-num-names}
\bibliography{iclr2022_conference}

\section*{Checklist}

\begin{enumerate}

\item For all authors...
\begin{enumerate}
  \item Do the main claims made in the abstract and introduction accurately reflect the paper's contributions and scope?
    \answerYes{}
  \item Did you describe the limitations of your work?
    \answerYes{See Assumptions and settings in Section \ref{sec:theo}.}
  \item Did you discuss any potential negative societal impacts of your work?
    \answerNA{}
  \item Have you read the ethics review guidelines and ensured that your paper conforms to them?
    \answerYes{}
\end{enumerate}

\item If you are including theoretical results...
\begin{enumerate}
  \item Did you state the full set of assumptions of all theoretical results?
    \answerYes{See Section \ref{sec:theo}.}
        \item Did you include complete proofs of all theoretical results?
    \answerYes{See Appendix \ref{app:1}}
\end{enumerate}

\item If you ran experiments...
\begin{enumerate}
  \item Did you include the code, data, and instructions needed to reproduce the main experimental results (either in the supplemental material or as a URL)?
    \answerYes{In the material.}
  \item Did you specify all the training details (e.g., data splits, hyperparameters, how they were chosen)?
    \answerYes{See Appendix \ref{appendix:expsetup}}
        \item Did you report error bars (e.g., with respect to the random seed after running experiments multiple times)?
    \answerNA{}
        \item Did you include the total amount of compute and the type of resources used (e.g., type of GPUs, internal cluster, or cloud provider)?
    \answerYes{The provider information Will be shown in Acknowledgement.}
\end{enumerate}

\item If you are using existing assets (e.g., code, data, models) or curating/releasing new assets...
\begin{enumerate}
  \item If your work uses existing assets, did you cite the creators?
    \answerYes{}
  \item Did you mention the license of the assets?
    \answerNo{The datatsets we used are well known.}
  \item Did you include any new assets either in the supplemental material or as a URL?
    \answerNo{}
  \item Did you discuss whether and how consent was obtained from people whose data you're using/curating?
    \answerNA{}
  \item Did you discuss whether the data you are using/curating contains personally identifiable information or offensive content?
    \answerNA{}
\end{enumerate}

\item If you used crowdsourcing or conducted research with human subjects...
\begin{enumerate}
  \item Did you include the full text of instructions given to participants and screenshots, if applicable?
    \answerNA{}
  \item Did you describe any potential participant risks, with links to Institutional Review Board (IRB) approvals, if applicable?
   \answerNA{}
  \item Did you include the estimated hourly wage paid to participants and the total amount spent on participant compensation?
    \answerNA{}
\end{enumerate}

\end{enumerate}


\appendix

\section{Detailed experimental setup}\label{appendix:expsetup}

For Fig. \ref{fig:flatness_cnn}(a), we use the FNN with size $784-1024-1024-10$. We add dropout layers behind the first and the second layers with dropout rate of 0.8 and 0.5, respectively. We train the network using default Adam optimizer \citep{kingma2015adam} with a learning rate of 0.0001.

For Fig. \ref{fig:flatness_cnn}(b), we use vgg-9 \citep{simonyan2014very} to compare the loss landscape flatness w/o dropout layers. For experiment with dropout layers, we add dropout layers after the pooling layers, the dropout rates of dropout layers are 0.8. Models are trained using GD with Nesterov momentum, training-size 2048 for 300 epochs. The learning rate is initialized at 0.1, and divided by a factor of 10 at epochs 150, 225 and 275. We only use the first 2048 examples for training to compromise with the computational burden. 

For Fig. \ref{fig:pca}(a, d), Fig. \ref{fig:Hessian}(a, d), Fig. \ref{fig:anisotropic}, we use the FNN with size $784-50-50-10$. We train the network using GD with the first 10,000 training data as the training set. We add a dropout layer behind the second layer. The dropout rate and learning rate are specified and unchanged in each experiment. We only consider the parameter matrix corresponding to the weight and the bias of the fully-connected layer between two hidden layers.

For Fig. \ref{fig:flatness_cnn}(c), Fig. \ref{fig:pca}(b, e), Fig. \ref{fig:Hessian}(b, e), we use ResNet-20 \citep{he2016deep} to compare the loss landscape flatness w/o dropout layers. For experiment with dropout layers, we add dropout layers after the convolutional layers, the dropout rates of dropout layers are $0.8$. We only consider the parameter matrix corresponding to the weight of the first convolutional layer of the first block of the ResNet-20. Models are trained using GD, training-size 50000 for 1200 epochs. The learning rate is initialized at 0.01. Since the Hessian calculation of ResNet takes much time, for the ResNet experiment, we only perform it at a specific dropout rate and learning rate.

For Fig. \ref{fig:flatness_cnn}(d), Fig. \ref{fig:pca}(c, f), Fig. \ref{fig:Hessian}(c, f), we use transformer \citep{vaswani2017attention} with $d_{model}=50, d_k=d_v=20, d_{ff}=256, h=4, N=3$, the meaning of the parameters is consistent with the original paper. In order to calculate the Hessian matrix and eigendecomposition more accurately and quickly, we reasonably reduce the number of network parameters. We only consider the parameter matrix corresponding to the weight of the fully-connected layer whose output is queries in the Multi-Head Attention layer of the first block of the decoder. For experiment with dropout layers, we apply dropout to the output of each sub-layer, before it is added to the sub-layer input and normalized. In addition, we apply dropout to the sums of the embeddings and the positional encodings in both the encoder and decoder stacks. The dropout rates of dropout layers are $0.9$. For the English-German translation problem, we use the cross-entropy loss with label smoothing trained by full-batch Adam based on the Multi30k dataset. The learning rate strategy is the same as that in the article. The warm up step is 4000 epochs, the training step is 10000 epochs. We only use the first 2048 examples for training to compromise with the computational burden.

\section{Derivations and Proofs for Main Paper}
\label{app:1}
\subsection{Proof of Theorem 1}

\begin{theorem*}[\textbf{Theorem 1: Hessian matrix with dropout regularization}]Based on the Setting \ref{set:1}-\ref{set:3} and Assumption \ref{ass:1}, the Hessian matrix of the loss function with respect to $\vf_{\vtheta,\veta}^{\mathrm{drop}}(\vx)$ can be written in the mean sense as:

\begin{equation*}
    H(\vtheta)=\Exp_{\veta}\nabla_{\vtheta}^{2}R_{S}^{\mathrm{drop}}=\frac{1}{n} \sum_{i=1}^{n}\left[ \nabla_{\vtheta} \vf_{\vtheta}\left(\vx_{i}\right) \nabla_{\vtheta}^{\T} \vf_{\vtheta}\left(\vx_{i}\right)+ \frac{1-p}{p} \sum_{j=1}^{m_{L-1}} \nabla_{\vtheta} \vf_{\vtheta}^{j}\left(\vx_{i}\right)  \nabla_{\vtheta}^{\T} \vf_{\vtheta}^{j}\left(\vx_{i}\right) \right].
\end{equation*}

\end{theorem*}

\begin{proof}

We first compute the Hessian matrix after taking expectation with respect to the dropout variable, 
\begin{equation}
    \Exp_{\veta}\nabla_{\vtheta}^{2}R_{S}^{\mathrm{drop}}(\vtheta,\veta)=\nabla_{\vtheta}^{2}R_{S}(\vtheta)+\frac{1-p}{2np}\sum_{i=1}^{n}\sum_{j=1}^{m_{L-1}}\nabla_{\vtheta}^{2}(\vf_{\vtheta}^{j}(\vx_{i}))^2.
    \label{equ:hess}
\end{equation}

The first and second terms on the RHS of the Equ.(\ref{equ:hess}) are as follows, 

\begin{equation*}
    \nabla_{\vtheta}^{2}R_{S}(\vtheta)=\frac{1}{n} \sum_{i=1}^{n}\left( \nabla_{\vtheta} \vf_{\vtheta}\left(\vx_{i}\right) \nabla_{\vtheta}^{\T} \vf_{\vtheta}\left(\vx_{i}\right)+(\vf_{\vtheta}\left(\vx_{i}\right)-\vy_{i})\cdot \nabla_{\vtheta}^{2}\vf_{\vtheta}\left(\vx_{i}\right) \right)
\end{equation*}
\begin{equation*}
    \frac{1-p}{2np}\sum_{i=1}^{n}\sum_{j=1}^{m_{L-1}}\nabla_{\vtheta}^{2}(\vf_{\vtheta}^{j}(\vx_{i}))^2 = \frac{1-p}{np} \sum_{i=1}^{n} \sum_{j=1}^{m_{L-1}}\left( \nabla_{\vtheta} \vf_{\vtheta}^{j}\left(\vx_{i}\right) \nabla_{\vtheta}^{\T} \vf_{\vtheta}^{j}\left(\vx_{i}\right)+\vf_{\vtheta}^{j}\left(\vx_{i}\right) \cdot\nabla_{\vtheta}^{2} \vf_{\vtheta}^{j}\left(\vx_{i}\right)\right). 
\end{equation*}
Note that for linear activate function, $\nabla_{\vtheta}^{2}\vf_{\vtheta}\left(\vx_{i}\right)=\nabla_{\vtheta}^{2} \vf_{\vtheta}^{j}\left(\vx_{i}\right)=\mzero,\ a.e. \ \forall i \in [n], \forall j \in [m]$, we have

\begin{equation*}
    \nabla_{\vtheta}^{2}R_{S}(\vtheta)=\frac{1}{n} \sum_{i=1}^{n} \nabla_{\vtheta} \vf_{\vtheta}\left(\vx_{i}\right) \nabla_{\vtheta}^{\T} \vf_{\vtheta}\left(\vx_{i}\right)
\end{equation*}
\begin{equation*}
    \frac{1-p}{2np}\sum_{i=1}^{n}\sum_{j=1}^{m_{L-1}}\nabla_{\vtheta}^{2}(\vf_{\vtheta}^{j}(\vx_{i}))^2 = \frac{1-p}{np} \sum_{i=1}^{n} \sum_{j=1}^{m_{L-1}} \nabla_{\vtheta} \vf_{\vtheta}^{j}\left(\vx_{i}\right)  \nabla_{\vtheta}^{\T} \vf_{\vtheta}^{j}\left(\vx_{i}\right). 
\end{equation*}

Thus the Equ.(\ref{equ:hess}) can be rewritten as
\begin{equation*}
    \Exp_{\veta}\nabla_{\vtheta}^{2}R_{S}^{\mathrm{drop}}(\vtheta,\veta)=\frac{1}{n} \sum_{i=1}^{n}\left( \nabla_{\vtheta} \vf_{\vtheta}\left(\vx_{i}\right) \nabla_{\vtheta}^{\T} \vf_{\vtheta}\left(\vx_{i}\right)+ \frac{1-p}{p} \sum_{j=1}^{m} \nabla_{\vtheta} \vf_{\vtheta}^{j}\left(\vx_{i}\right)  \nabla_{\vtheta}^{\T} \vf_{\vtheta}^{j}\left(\vx_{i}\right) \right).
\end{equation*}
\end{proof}

\subsection{Proof of Theorem 2}

\begin{theorem*}[\textbf{Theorem 2: Covariance matrix with dropout regularization}]Based on the Setting \ref{set:1}-\ref{set:3} and Assumption \ref{ass:1}, the covariance matrix of the loss function under the randomness of dropout variable $\veta$ and data $\vx$ can be written as:

\begin{equation*}
    \begin{aligned}
    \Sigma^{\mathrm{drop}}_{\vtheta}&=\frac{1}{n}\sum_{i=1}^{n} \left[ a_i \nabla_{\vtheta}\vf_{\vtheta}(\vx_i)\nabla_{\vtheta}^{\T}\vf_{\vtheta}(\vx_i)+ b_{i} \frac{1-p}{p}\sum_{j=1}^{m_{L-1}} \nabla_{\vtheta}\vf_{\vtheta}^{j}(\vx_i)\nabla_{\vtheta}^{\T}\vf_{\vtheta}^{j}(\vx_i)\right],
    \end{aligned}
\end{equation*}
where $a_i=2\Exp_{\veta}\ell(\vf_{\vtheta,\veta}^{\mathrm{drop}}(\vx_i),\vy_{i})$, $b_{i}=2\ell(\vf_{\vtheta}(\vx_i),\vy_{i})$.

\end{theorem*}
\begin{proof}
For simplicity, we approximate the loss function through Taylor expansion, which is also used in \cite{wei2020implicit},  
\begin{equation*}
    \ell(\vf_{\vtheta,\veta}^{\mathrm{drop}}(\vx_i),\vy_{i}) \approx \ell(\vf_{\vtheta}(\vx_i),\vy_{i})+(\vf_{\vtheta}(\vx_i)-\vy_{i})\sum_{j=1}^{m_{L-1}}\mW^{[L]}_{j} (\veta)_{j} \vf^{[L-1]}_{\vtheta, j}(\vx_i),
\end{equation*}
where $\mW^{[L]}_{j} \in  \sR^{m_L}$ is the $j$th column of $\mW^{[L]}$, and $\vf^{[L-1]}_{\vtheta, j}(\vx_i)$ is the $j$th element of vector $\vf^{[L-1]}_{\vtheta}(\vx_i)$. The covariance matrix obtained using SGD under dropout regularization is
\begin{equation*}
\begin{aligned}
    \Sigma^{\mathrm{drop}}_{\vtheta}
    &=\frac{1}{n}\sum_{i=1}^{n}\Exp_{\veta}\nabla_{\vtheta}\ell(\vf_{\vtheta,\veta}^{\mathrm{drop}}(\vx_i),\vy_{i}) \nabla_{\vtheta}^{\T}\ell(\vf_{\vtheta,\veta}^{\mathrm{drop}}(\vx_i),\vy_{i}) - \nabla_{\vtheta} \Exp_{\veta}R_{S}^{\mathrm{drop}}(\vtheta,\veta) \nabla_{\vtheta}^{\T} \Exp_{\veta}R_{S}^{\mathrm{drop}}(\vtheta,\veta)\\
    &\approx \frac{1}{n}\sum_{i=1}^{n}\Exp_{\veta}\nabla_{\vtheta}\ell(\vf_{\vtheta,\veta}^{\mathrm{drop}}(\vx_i),\vy_{i}) \nabla_{\vtheta}^{\T}\ell(\vf_{\vtheta,\veta}^{\mathrm{drop}}(\vx_i),\vy_{i}).
\end{aligned}
\end{equation*}
Combining the properties of the dropout variable $\veta$, we have, 
\begin{equation}
\begin{aligned}
    \Sigma^{\mathrm{drop}}_{\vtheta}&=\frac{1}{n}\sum_{i=1}^{n}\nabla_{\vtheta}\ell(\vf_{\vtheta}(\vx_i),\vy_{i}) \nabla_{\vtheta}^{\T}\ell(\vf_{\vtheta}(\vx_i),\vy_{i}) \\
    &+\frac{1}{n}\sum_{i=1}^{n} \Exp_{\veta} \left(\sum_{j=1}^{m_{L-1}}(\veta)_{j}\nabla_{\vtheta}(\vf_{\vtheta}^{j}(\vx_i)\vepsilon_{\vtheta,i})\sum_{j=1}^{m_{L-1}}(\veta)_{j}\nabla_{\vtheta}^{\T}(\vf_{\vtheta}^{j}(\vx_i)\vepsilon_{\vtheta,i}) \right) \\
    &=\frac{1}{n}\sum_{i=1}^{n}\left(\nabla_{\vtheta}\ell(\vf_{\vtheta}(\vx_i),\vy_{i}) \nabla_{\vtheta}^{\T}\ell(\vf_{\vtheta}(\vx_i),\vy_{i})+\frac{1-p}{p} \sum_{j=1}^{m_{L-1}}\nabla_{\vtheta}(\vf_{\vtheta}^{j}(\vx_i)\vepsilon_{\vtheta,i})\nabla_{\vtheta}^{\T}(\vf_{\vtheta}^{j}(\vx_i)\vepsilon_{\vtheta,i}) \right)\\
    &:= \frac{1}{n}\sum_{i=1}^{n}\left(\Sigma^{\mathrm{drop}}_{\vtheta,1}(\vx_{i},\vy_{i})+\frac{1-p}{p}\Sigma^{\mathrm{drop}}_{\vtheta,2}(\vx_{i},\vy_{i})\right).
\end{aligned}    
\label{equ:cov}
\end{equation}

We calculate the two terms on the RHS of the Equ.(\ref{equ:cov}) separately:

\begin{equation*}
    \Sigma^{\mathrm{drop}}_{\vtheta,1}(\vx_{i},\vy_{i})=(\vepsilon_{\vtheta}(\vx_i))^2 \cdot \nabla_{\vtheta}\vf_{\vtheta}(\vx_i)\nabla_{\vtheta}^{\T}\vf_{\vtheta}(\vx_i),
\end{equation*}

\begin{equation*}
\begin{aligned}
    \Sigma^{\mathrm{drop}}_{\vtheta,2}(\vx_{i},\vy_{i})&=(\vepsilon_{\vtheta,i})^2\sum_{j=1}^{m_{L-1}}\nabla_{\vtheta}\vf_{\vtheta}^{j}(\vx_i)\nabla_{\vtheta}^{\T}\vf_{\vtheta}^{j}(\vx_i)+\nabla_{\vtheta}\vf_{\vtheta}(\vx_i)\nabla_{\vtheta}^{\T}\vf_{\vtheta}(\vx_i)\sum_{j=1}^{m_{L-1}}(\vf_{\vtheta}^{j}(\vx_i))^2\\
    &+2\sum_{j=1}^{m_{L-1}}\vepsilon_{\vtheta,i}\vf_{\vtheta}^{j}(\vx_i)\cdot\nabla_{\vtheta}\vepsilon_{\vtheta,i}\nabla_{\vtheta}\vf_{\vtheta}^{j}(\vx_i)\\
    &=(\vepsilon_{\vtheta,i})^2\sum_{j=1}^{m_{L-1}}\nabla_{\vtheta}\vf_{\vtheta}^{j}(\vx_i)\nabla_{\vtheta}^{\T}\vf_{\vtheta}^{j}(\vx_i)+\nabla_{\vtheta}\vf_{\vtheta}(\vx_i)\nabla_{\vtheta}^{\T}\vf_{\vtheta}(\vx_i)\sum_{j=1}^{m_{L-1}}(\vf_{\vtheta}^{j}(\vx_i))^2\\
    &+\frac{1}{2}\sum_{j=1}^{m_{L-1}}\nabla_{\vtheta}(\vepsilon_{\vtheta,i})^2\nabla_{\vtheta}(\vf_{\vtheta}^{j}(\vx_i))^2.
\end{aligned}
\end{equation*}
Under the assumption that $\nabla_{\vtheta}(\vepsilon_{\vtheta,i})^2=2\cdot \nabla_{\vtheta}\ell(\vf_{\vtheta}(\vx_i),\vy_{i})=\mzero$, $\forall i \in [n]$, we have 
\begin{equation*}
    \Sigma^{\mathrm{drop}}_{\vtheta,2}(\vx_{i},\vy_{i})=(\vepsilon_{\vtheta,i})^2\sum_{j=1}^{m_{L-1}}\nabla_{\vtheta}\vf_{\vtheta}^{j}(\vx_i)\nabla_{\vtheta}^{\T}\vf_{\vtheta}^{j}(\vx_i)+\nabla_{\vtheta}\vf_{\vtheta}(\vx_i)\nabla_{\vtheta}^{\T}\vf_{\vtheta}(\vx_i)\sum_{j=1}^{m_{L-1}}(\vf_{\vtheta}^{j}(\vx_i))^2.
\end{equation*}
Thus the Equ.(\ref{equ:cov}) can be rewritten as
\begin{equation*}
\begin{aligned}
    \Sigma^{\mathrm{drop}}_{\vtheta}&=\frac{1}{n}\sum_{i=1}^{n}\nabla_{\vtheta}\vf_{\vtheta}(\vx_i)\nabla_{\vtheta}^{\T}\vf_{\vtheta}(\vx_i)\left((\vepsilon_{\vtheta}(\vx_i))^2+ \frac{1-p}{p}\sum_{j=1}^{m_{L-1}}(\vf_{\vtheta}^{j}(\vx_i))^2\right)\\
    &+\frac{1-p}{np}\sum_{i=1}^{n}\sum_{j=1}^{m_{L-1}}(\vepsilon_{\vtheta,i})^2\cdot\nabla_{\vtheta}\vf_{\vtheta}^{j}(\vx_i)\nabla_{\vtheta}^{\T}\vf_{\vtheta}^{j}(\vx_i).
\end{aligned}
\end{equation*}
Note that 
\begin{equation*}
    (\vepsilon_{\vtheta}(\vx_i))^2+ \frac{1-p}{p}\sum_{j=1}^{m_{L-1}}(\vf_{\vtheta}^{j}(\vx_i))^2=\Exp_{\veta}2\ell(\vf_{\vtheta,\veta}^{\mathrm{drop}}(\vx_i),\vy_{i}), 
\end{equation*}
we have
\begin{equation*}
    \begin{aligned}
    \Sigma^{\mathrm{drop}}_{\vtheta}&=\frac{2}{n}\sum_{i=1}^{n}\Exp_{\veta}\ell(\vf_{\vtheta,\veta}^{\mathrm{drop}}(\vx_i),\vy_{i})\cdot \nabla_{\vtheta}\vf_{\vtheta}(\vx_i)\nabla_{\vtheta}^{\T}\vf_{\vtheta}(\vx_i)\\
    &+\frac{2(1-p)}{np}\sum_{i=1}^{n}\sum_{j=1}^{m_{L-1}}(\ell(\vf_{\vtheta}(\vx_i),\vy_{i}))\cdot\nabla_{\vtheta}\vf_{\vtheta}^{j}(\vx_i)\nabla_{\vtheta}^{\T}\vf_{\vtheta}^{j}(\vx_i).
    \end{aligned}
\end{equation*}
\end{proof}

\subsection{Proof of Proposition 1}

\begin{proposition*}[\textbf{Proposition 1}]
Based on the Setting \ref{set:1}-\ref{set:3} and Assumption \ref{ass:1}, we further restrict the problem to a binary classification problem, i.e. $\vy_{i} \in \{0,1\},\ \forall i \in [n]$, and assume the model output $\vf_{\vtheta}(\vx_i) \in [\delta, 1-\delta] $ (we can limit the network output using a threshold activation function), where $\delta$ is a small positive constant, then we have:

(i) $\Sigma^{\mathrm{drop}}_{\vtheta} \succeq \delta^{2} H(\vtheta)$, almost everywhere in $\sR^{M}$, $M$ is the dimension of $\vtheta$; 

(ii) For any $\epsilon>0$, and a network parameter $\vtheta\in\Omega=\{\vtheta: \Exp_{\veta}\ell(\vf_{\vtheta,\veta}^{\mathrm{drop}}(\vx_i),\vy_{i}) \leq  \frac{ (\delta+\epsilon)^{2}}{2},\ell(\vf_{\vtheta}(\vx_i),\vy_{i}) \leq  \frac{ (\delta+\epsilon)^{2}}{2},  \forall i \in [n]\}$, we have $\Sigma^{\mathrm{drop}}_{\vtheta} \preceq  (\delta+\epsilon)^{2}  H(\vtheta)$ almost everywhere in $\Omega$. 
\end{proposition*}

\begin{proof}
    The properties (i)–(ii) are direct consequences of Thm. \ref{thm:1}, \ref{thm:2}. 
\end{proof}

\end{document}